\def\year{2022}\relax
\documentclass[letterpaper]{article} 
\usepackage{aaai22}  
\usepackage{times}  
\usepackage{helvet}  
\usepackage{courier}  
\usepackage[hyphens]{url}  
\usepackage{graphicx} 
\usepackage{natbib}  
\usepackage{caption} 
\DeclareCaptionStyle{ruled}{labelfont=normalfont,labelsep=colon,strut=off} 
\usepackage{algorithm}
\usepackage{algorithmic}

\usepackage{newfloat}
\usepackage{listings}
\floatstyle{ruled}
\newfloat{listing}{tb}{lst}{}
\floatname{listing}{Listing}
\usepackage{booktabs}
\usepackage{amsmath}
\usepackage{amssymb}

\title{Evaluating Deep Taylor Decomposition for Reliability Assessment in the Wild}
\author{
    Stephanie Brandl, 
    Daniel Hershcovich, 
    Anders Søgaard 
}
\affiliations {
    University of Copenhagen\\
    \{brandl, dh, soegaard\}@di.ku.dk
}

\begin{document}
\maketitle
\begin{abstract}
We argue that we need to evaluate model interpretability methods 'in the wild', i.e., in situations where professionals make critical decisions, and models can potentially assist them. We present an in-the-wild evaluation of token attribution based on Deep Taylor Decomposition, with professional journalists performing reliability assessments. We find that using this method in conjunction with RoBERTa-Large, fine-tuned on the Gossip Corpus, led to faster and better human decision-making, as well as a more critical attitude toward news sources among the journalists. We present a comparison of human and model rationales, as well as a qualitative analysis of the journalists' experiences with machine-in-the-loop decision making.
\end{abstract}
\vspace{3mm}
\section{Introduction}

Deep neural NLP models increasingly assist humans in making documents easier to find and analyze. Generally, models are used for either {\em batch processing}, e.g., summarizing or indexing documents, or for {\em online decision support}, e.g., flagging probable fraud or toxic speech. Deep neural models are often thought of as black boxes, unable to provide rationales for their decisions, but recently, many methods have been developed for post-hoc interpretation of deep neural model predictions \cite{guidotti2018survey,ancona2018towards,poerner-etal-2018-evaluating,atanasova-etal-2020-diagnostic}. Most such methods provide rationales in the form of input token attributions. While alternatives exist \cite{soegaard2021explainable}, we evaluate attributions below. 

Rationales in the form of input token attributions have traditionally been evaluated automatically or through comparison with gold-standard, human rationales. Automatic evaluation methods look at the impact of removing (or keeping only) the tokens that are attributed most relevance or influence, while comparisons with human rationales typically evaluate predicted rationales in terms of matching metrics such as token-level $F_1$. 

\begin{figure}
    \centering
    \hfill \includegraphics[width=\columnwidth]{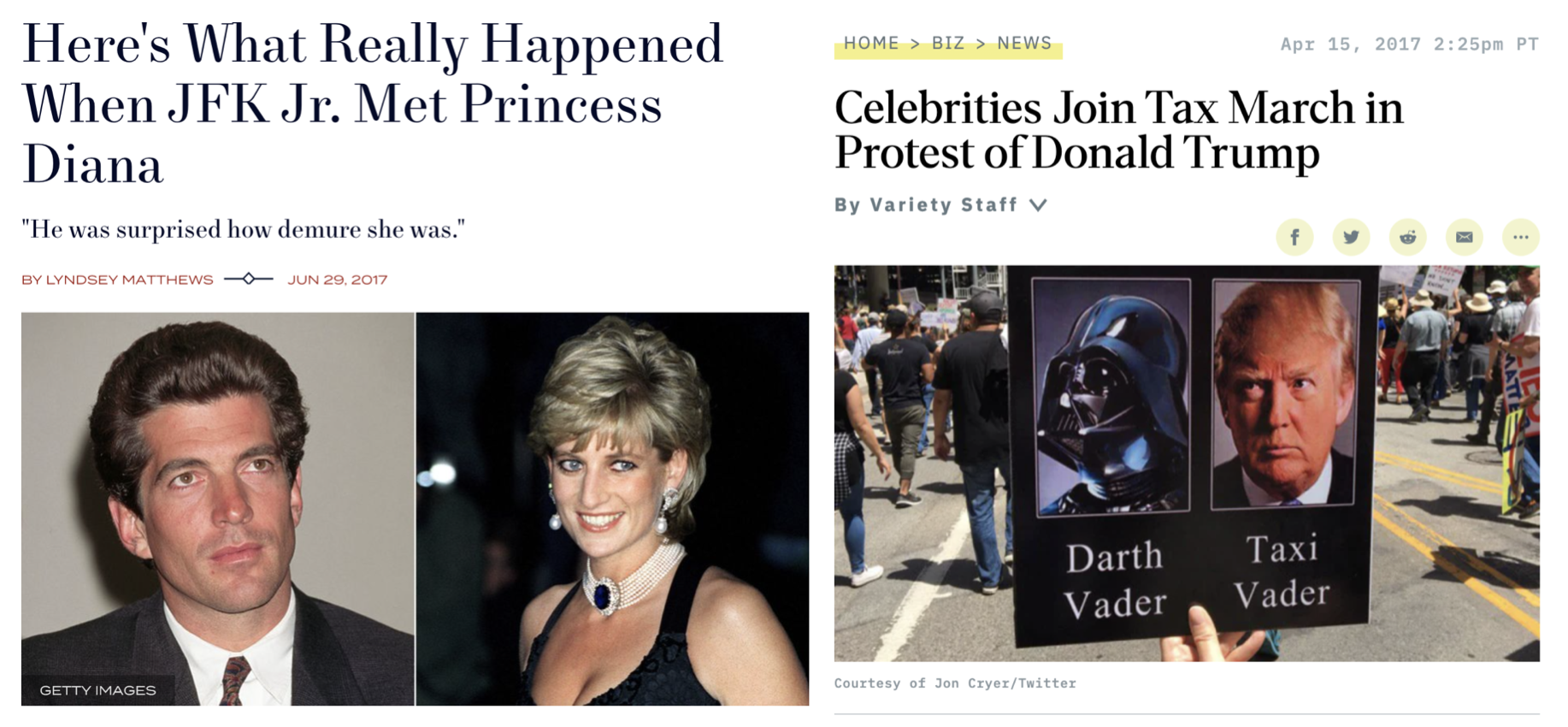}
    \caption{Celebrity news stories, the text of which (both longer than shown) are included in the GossipCop corpus. These examples are labeled {\em unreliable}. We ask professional journalists to relabel the texts, with or without the help of a RoBERTa-based document classifier and Deep Taylor Decomposition rationales.}
    \label{fig:example}\end{figure}
    
\paragraph{Related work} \citet{horne2019rating}  evaluated if machine learning with and without feature attribution rationales can help {\em lay people} assess the reliability of claims. We follow their data collection set-up in most respects, but not their experimental protocol. We also rely on a different source of data.  \citet{horne2019rating} found a positive effect of model rationales on human decision making, but evaluated this only in terms of mean ratings for reliable and unrealiable articles, leaving it open whether human decision-making was more accurate at the article level. They found no positive effect of showing just model confidences. The most important difference between their study and ours, however, is that we evaluate the usefulness of reliability detection with feature attribution rationales on {\em professional experts} (journalists). 

More recently, \citet{mohseni2021machine} evaluated the effect of model rationales (using self-explanatory models) across four fake news detection models, also with lay people. They found that using rationales is not significantly better than showing only model confidence, and generally found the effectiveness of explanations to vary across architectures. Our results align better with  \citet{horne2019rating} than with \citet{mohseni2021machine}, but provide a more nuanced picture and for a population of end users, who are trained to do reliability assessment, and who have a real need to perform this activity on a day-to-day basis. 

In older work, \citet{feng2019can} evaluated the usefulness of model rationales in the context of human question answering. They recruited trivia experts and novices to play
QuizBowl with a computer as their teammate, with models returning both top-$k$ predictions and rationales. Rationales was provided both in terms of feature attributions and training examples. Their work is also limited in several respects, relying on an inherently interpretable linear question answering model and therefore, in effect, not evaluating interpretability methods, but simply the usefulness of an interpretable model. Their task is also different than ours in having an open-ended output space. This means that simply listing top-10 candidate answers can contribute to human recollection of answers, an effect they exploit, but which we cannot for a binary classification task. \citet{feng2019can} find that only novices generally benefit from model rationales. Finally, QuizBowl is arguably not a task where interpretability is needed -- in the sense that this is a necessary feature for decision support in reliability assessment to avoid the spread of misinformation. In sum, their work leaves open a) whether interpretability methods are useful when applied to deep neural models, b) whether  interpretability methods can be useful to experts, and c) whether interpretability methods can be useful in situations where interpretability is critical. While \citet{horne2019rating} and \citet{mohseni2021machine} addressed a) and b), our work provides partial answers to all these three questions. 

Other work that addresses human-in-the-loop evaluation of interpretability for deep neural models (a) includes \citet{gonzales2020reverse} and \citet{gonzalez-etal-2021-explanations}, but both evaluate interpretability methods with lay people and on non-critical tasks, ignoring (b) and (c). Attempts to evaluate interpretability methods for experts performing critical tasks, have, to the best of our knowledge, been limited to automatic evaluation or evaluation against gold-standard human rationales. 

\paragraph{Contributions} In sum, our main contribution compared to previous work is that we evaluate feature attribution rationales for reliability assessment on professional journalists. While misinformation spread in social media is devastating, misinformation accelerates when authoritative platforms, including traditional media, pass on unreliable stories \cite{soares2021mainstream}. As a way of evaluating interpretability methods, professional journalists present a much higher bar. Journalists are already trained to assess the reliability of news stories, and predictions and rationales have to be of very high quality to assist, rather than bias, such professional end users. In addition, we also evaluate a novel, state-of-the-art interpretability algorithm (Deep Taylor Decomposition for deep neural networks) on a new source of data, namely a corpus of celebrity gossip stories, a domain which is arguably a greater source of misinformation than any other domain and has been estimated to account for up to half of all online misinformation \cite{acerbi2019cognitive}. Finally, we consider the impact of the demographics and experience of the professional journalists on the usefulness of feature attribution.

\begin{figure}[h]
    \centering
    \includegraphics[width=.5\textwidth]{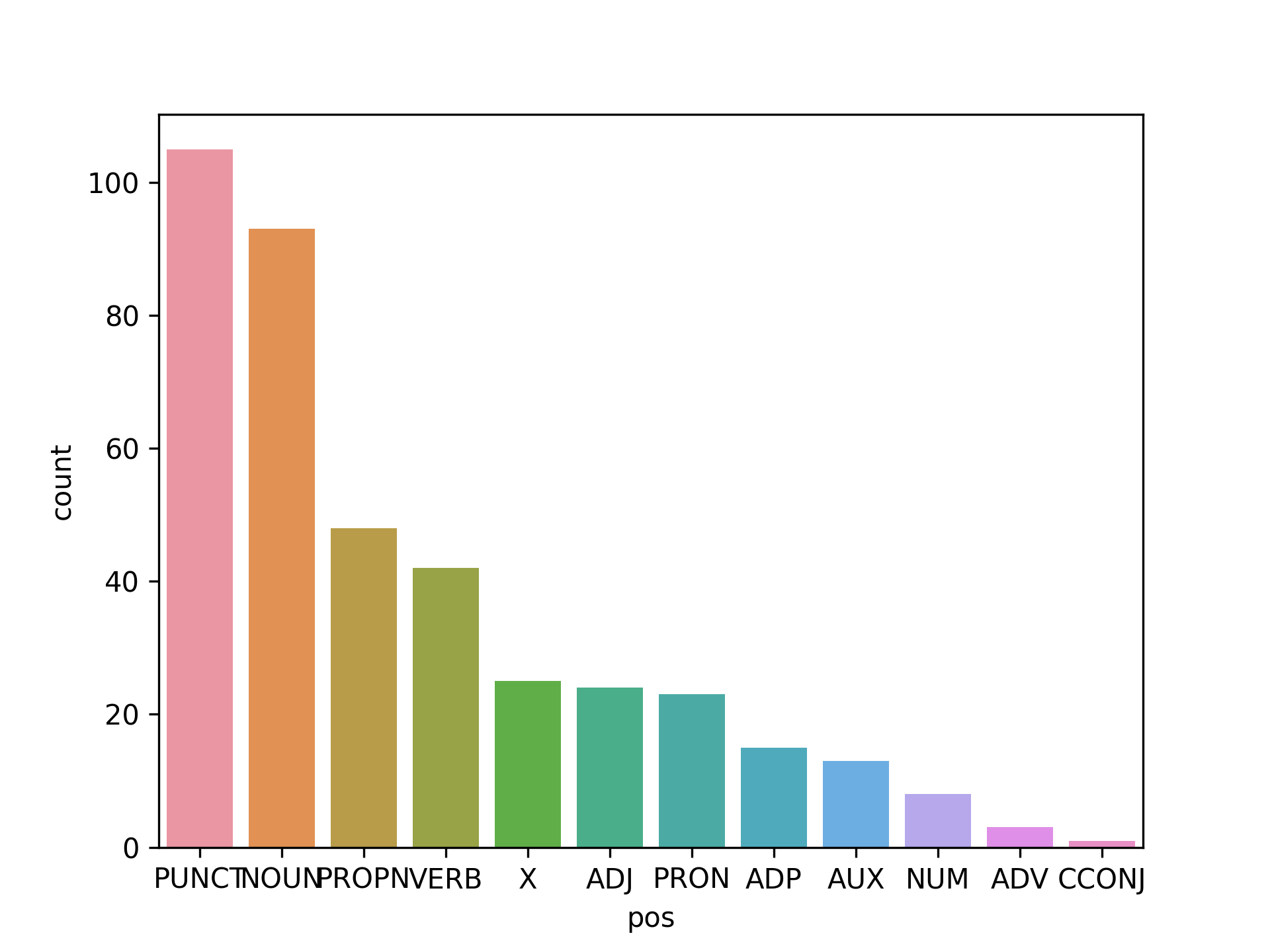}
    \caption{POS distribution of most highlighted words}
    \label{fig:pos}
\end{figure}

\begin{figure*}[h]
    \centering
    \includegraphics[width=6in]{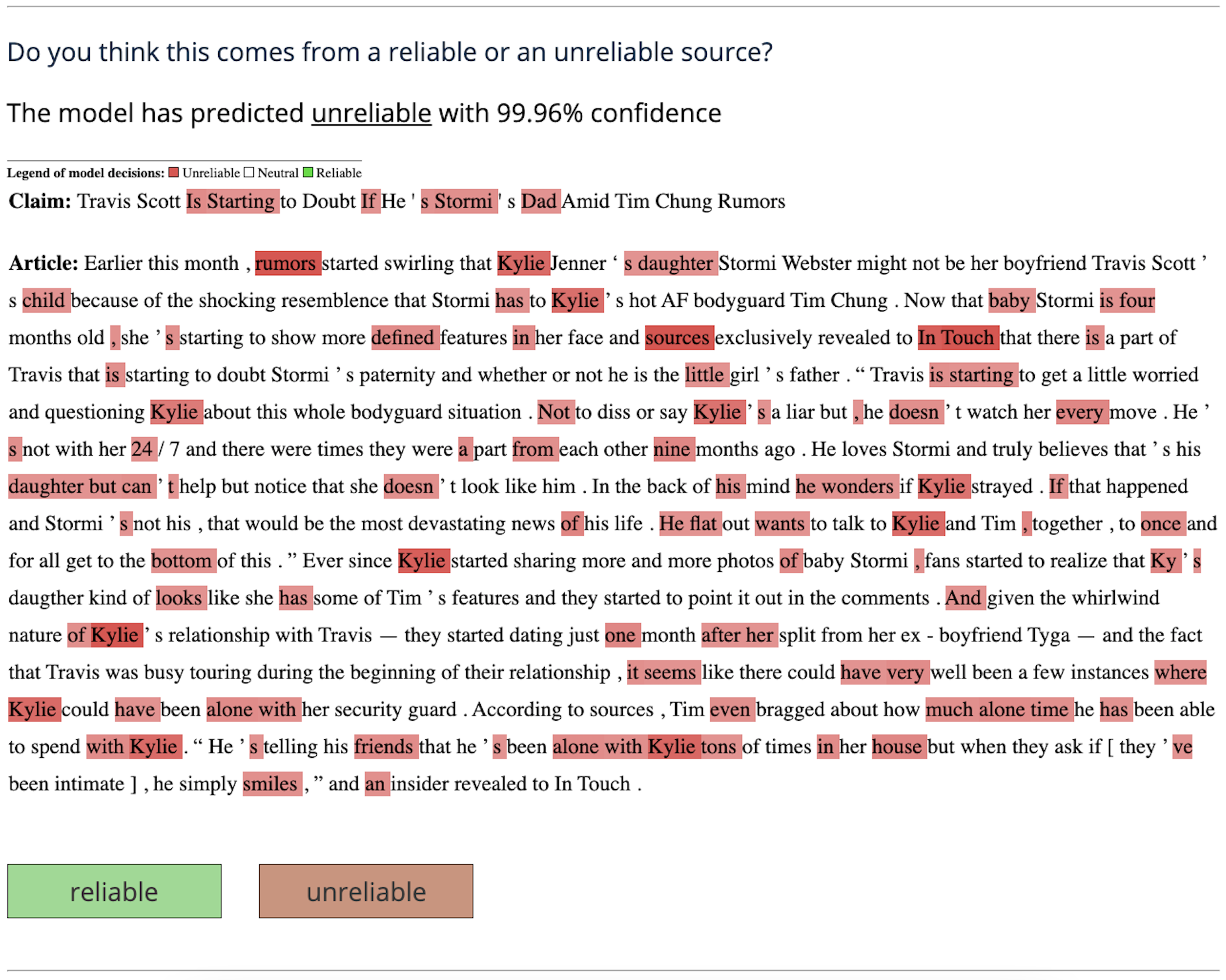}
    \caption{The interface used in our human reliability assessment with professional journalists receiving model prediction (with confidence) and rationales as feedback.}
    \label{fig:ui}
\end{figure*}

\begin{table}[h]
    \centering
    \begin{tabular}{lcc}
         \toprule 
         &Time&Error\\
         \midrule 
         Text&{\bf 45.8}&0.31\\
         +  Confidence&49.2&0.25\\
         \midrule 
         +  Confidence + Feat.attr.&{48.6}&{\bf 0.22}\\
         \bottomrule
    \end{tabular}
    \caption{Deep Taylor Decomposition (last row) enables faster (Time) and better (Error) reliability assessment, even for professional journalists.}
    \label{tab:results}
\end{table}

\begin{figure}[h!]
    \centering
    \includegraphics[width=.85\columnwidth]{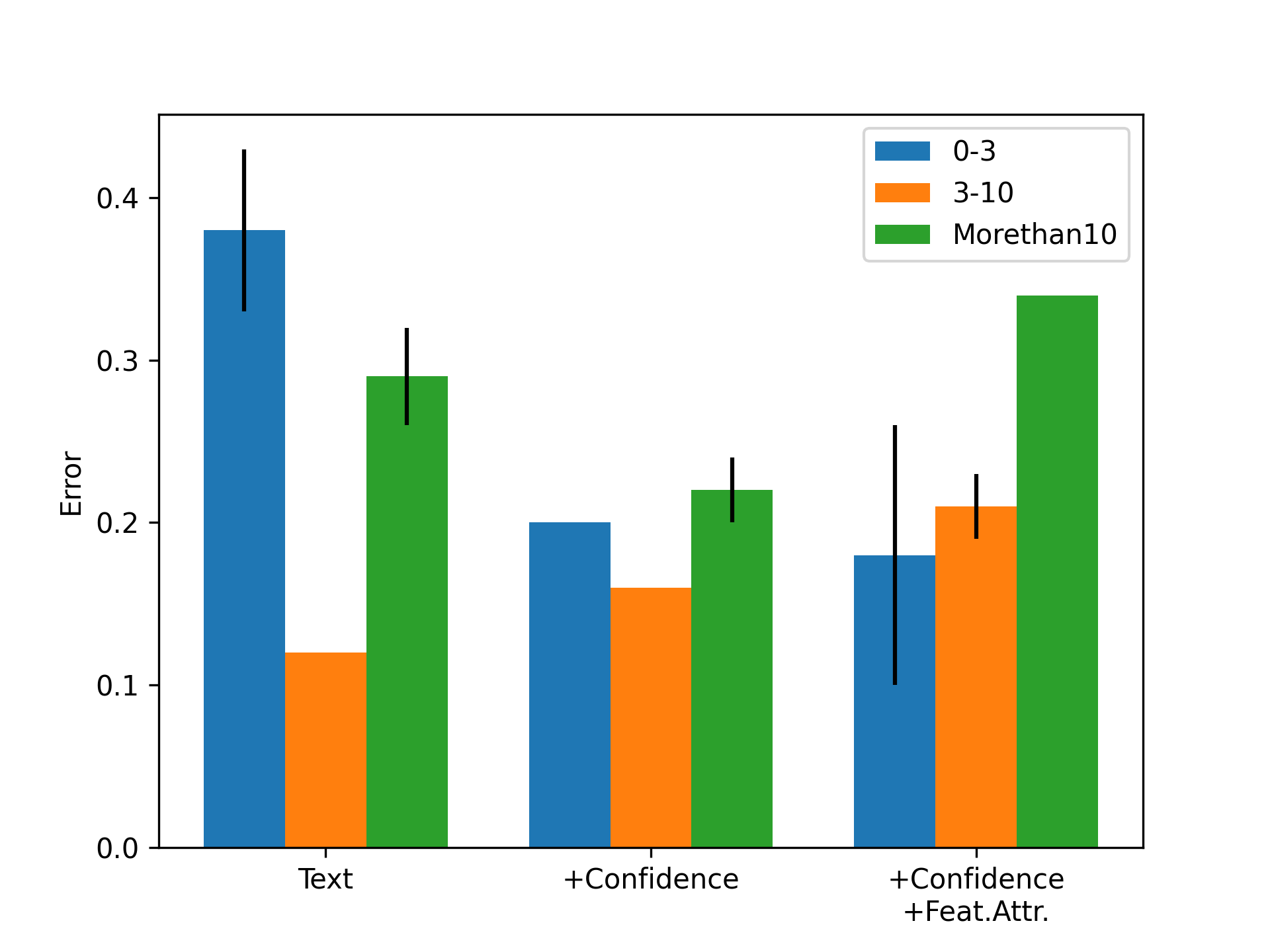}
        \includegraphics[width=.85\columnwidth]{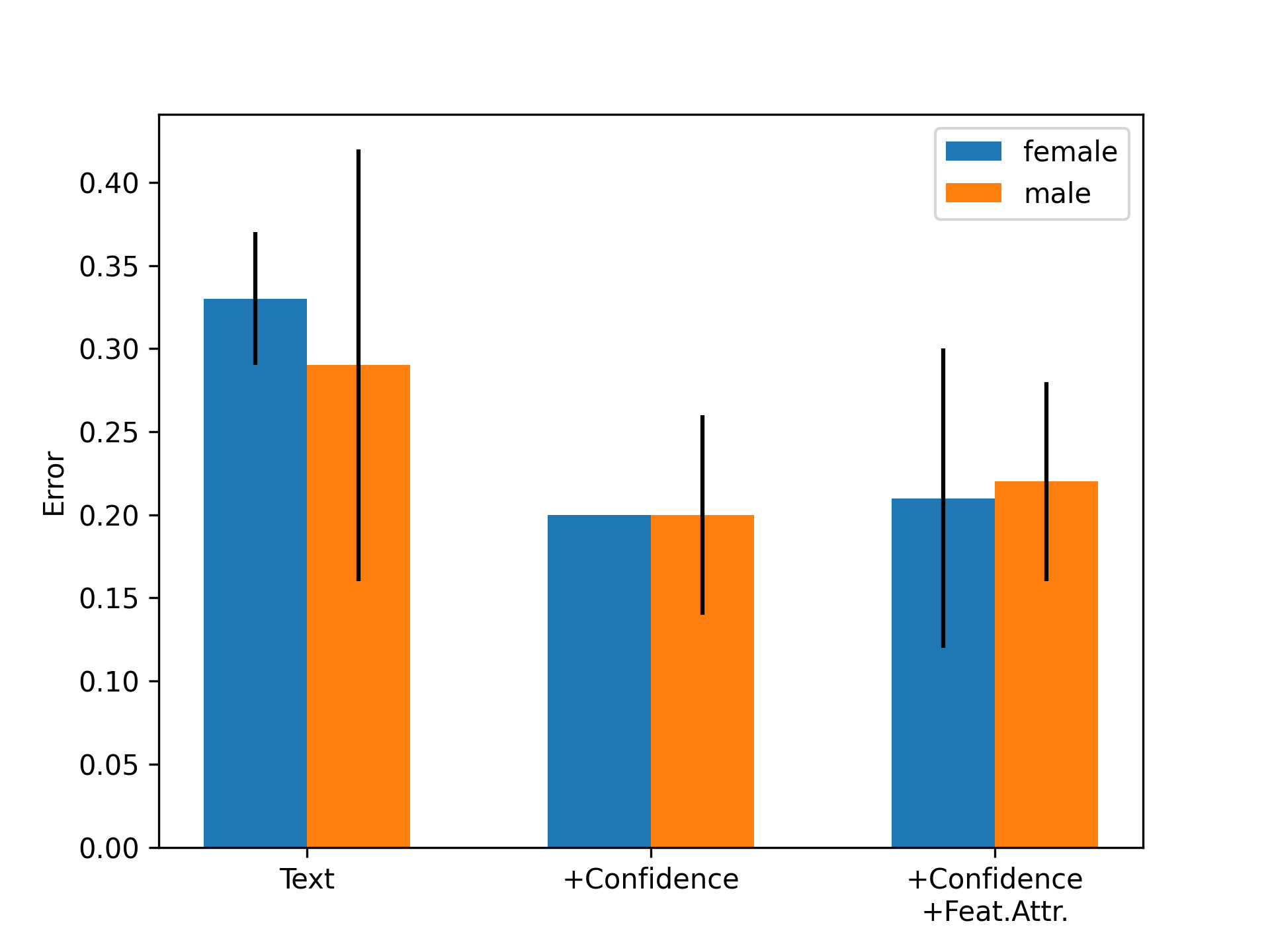}
            \includegraphics[width=.85\columnwidth]{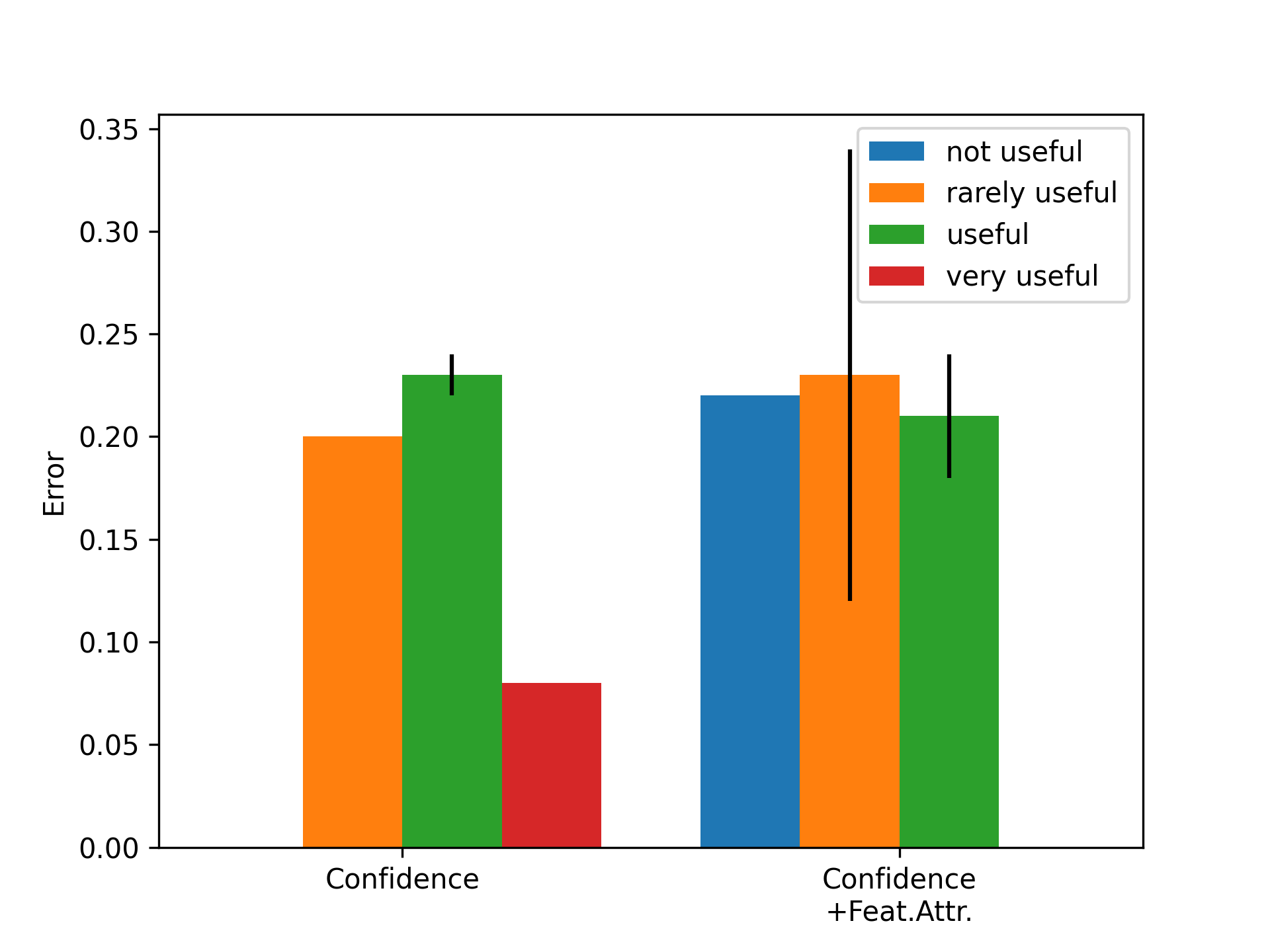}
    \caption{Error rate for all 3 conditions with journalists grouped by experience (top), gender (middle), and 
    by how useful they rated the model feedback (right). Error bars are standard deviation across bootstrapped samples.}
    \label{fig:experience}
\end{figure}

\section{Experimental Setup}

\paragraph{Participants} We recruited 25 professional journalists (6 female, 17 male, 2 without response) to participate in our experiment. All journalists were employed at the time of the experiment in the same media house, working for a tabloid newspaper, and had in this capacity daily contact with reliability assessment of celebrity-related news. The participants were compensated through salary. 

\paragraph{Data} We sampled 80 articles for our experiments from the GossipCop section of the FakeNewsNet \cite{shu2019fakenewsnet}.\footnote{\url{github.com/KaiDMML/FakeNewsNet}} The articles were split in 2 batches of 40 articles, half of which our model (see below) had classified incorrectly. In each condition, each article was seen by a minimum of three journalists. Gossip as a genre has the advantage that reliability assessment here should be relatively unbiased by domain experience and political alignment, evaluating instead the plausibility of a reported state of affairs. We provide an example in Table~\ref{fig:example}. 

\paragraph{Model} We train a binary document classifier by fine-tuning RoBERTa-Large \cite{liu2019roberta} on the rest of the GossipCop data. RoBERTa-Large -- a pre-trained English language model based on Transformer blocks \cite{NIPS2017_3f5ee243} -- is augmented with a simple classification architecture to calculate cross-entropy. Our architecture is completely standard and applies drop-out to the Transformer's last layer hidden state. The regularized output is then fed into a single fully-connected layer and a softmax activation function to scale logits to probabilities. We use Deep Taylor Decomposition \cite{chefer2021transformer} to assign feature attribution scores to input words. See code for details. Deep Taylor Decomposition is non-trivial in Transformer models because propagation involves attention layers and skip connections, but \citet{chefer2021transformer}~present a novel, consistent method for computing attribution scores across such layers and connections.

\paragraph{Predictive words} Deep Taylor Decomposition assigns feature attribution scores to input words. Reviewing aggregate statistics across the examples used in our experiments, we see the POS distribution for the most highlighted words in Figure~\ref{fig:pos}. We see our model is very sensitive to punctuation, e.g., misplaced commas, and nouns, e.g., first names without last names. Several of the journalists, when asked for signs of unreliability, mentioned poor grammar as the main indicator outside of metadata.

\paragraph{Stimuli presentation} The feature attribution scores were displayed to the journalists in one of three conditions; see Figure~\ref{fig:ui} to see the user interface. The interfaces for the other conditions were identical, except without color highlights and/or model predictions and confidence scores. 

\paragraph{Metrics} We evaluate the time used by professional journalists to assess the reliability of the news articles in terms of average wall-clock time in seconds. We evaluate their ability to assess reliability by evaluating their accuracy (1-0 loss) compared to the GossipCop gold-standard annotations.

\section{Results}

Our main results are listed in Table~\ref{tab:results}. We see that the time journalists take to assess reliability is not significantly impacted by the model feedback. We see a 3.5s increase on average when introducing model predictions and confidence scores, but on the other hand, the highlighting seems to increase reading speed a bit, leading to 0.6s faster reading times than in the setup without feature attributions. 

More importantly, human error rate is substantially lower for professional journalists when they receive model feedback. In the baseline condition of {\em no}~model feedback, they are only able to correctly estimate the reliability of two in three documents, whereas with model confidence scores, their accuracy is three in four, i.e., the error rate is 0.25. Model rationales in the form of feature attribution further decreases the error rate to 0.22.

\section{Analysis}

As part of our experiment, we also asked the journalists to volunteer information about their seniority, gender, and how useful they found the model feedback. We use this additional survey data to perform a more fine-grained analysis of the usefulness of the feature attribution scores of Deep Taylor Decomposition for the task of reliability assessment. 

\paragraph{Experience} The 25 professional journalists in our experiment all work for the same media house, but some are more senior than others. The usefulness of feature attribution has previously been said to correlate with level of experience \cite{feng2019can}. Our data corroborates this claim (see Figure~\ref{fig:experience}, left chart). In fact, it turns out the positive effect of interpretability is entirely with journalists with less than three years of experience. Model feedback, in contrast, seems to hurt the reliability assessments of more experienced journalists. 

\paragraph{Demographics} We also considered whether reported gender of the participants had an effect on the usefulness of model feedback, i.e., are men or women more prone to model suggestions? The positive effect of interpretability methods seems insensitive to the gender of the journalists. See Figure~\ref{fig:experience} (middle) for numbers.\footnote{23/25 reported they identified as either male or female.}


\paragraph{Usefulness} Finally, we examined whether the perceived usefulness of the model feedback by the journalists correlated with their error. Mostly, it did not, but the small group of journalists who found the model feedback {\em very useful} had very low error rates. Most journalists rated model predictions as {\em useful} or {\em very useful}, whereas most journalists rated model predictions and rationales as {\em not useful} or {\em rarely useful}. This is remarkable in light of the better performance of journalists relying on both model predictions and rationales. It seems that {\bf showing model predictions and rationales to professional journalists improves their reliability assessments, but hurts their perception of the usefulness of having a model in the loop}.\footnote{This seems paradoxical, but there are many similar paradoxes in human decision-making, e.g., \citet{doi:https://doi.org/10.1002/9780470400531.eorms0636}.}


\paragraph{Qualitative feedback} The journalists were also asked to provide qualtiative feedback on their experience with model-supported reliability assessment. Example feedback ranged from {\em the model and the words marked worked fine for me}~to {\em semed random at times}. Several journalists explicitly referred to the model as a source of bias, acknowledging its influence on their assessments. 

\section{Conclusions} We evaluated feature attributions of Deep Taylor Decomposition applied to a deep neural document classifier based on fine-tuned RoBERTa-Large representations, in a model-assisted human decision-making experiment with 25 professional journalists performing in-the-wild reliability assessments of English celebrity news stories. We found feature attribution to have a positive effect on the ability of journalists to assist the reliability of these stories. We also saw, however, that this effect was only with journalists with less than three years of experience. Interestingly, while the feature attribution scores had a positive net effect on reliability assessment, a setup where only model predictions and confidence scores were provided, was perceived of as more useful by the participants. We make the data and code involved in our experiments publicly available for replication and future work.\footnote{\url{https://github.com/coastalcph/reliability-wild}}

\section*{Acknowledgements} This  work  was partially funded by the Platform Intelligence in News project, which is supported by Innovation Fund Denmark via the Grand Solutions program. Thanks to Ana Valeria Gonzalez for help with designing our experiments. 

\bibliography{References}
\end{document}